\pdfoutput=1
\documentclass[11pt]{article}
\usepackage{acl}
\usepackage{times}
\usepackage{latexsym}
\usepackage[T1]{fontenc}
\usepackage{microtype}
\usepackage{inconsolata}
\usepackage{graphicx}
\usepackage{hyperref}
\usepackage{amsmath}
\usepackage{algorithm}
\usepackage{algorithmicx}
\usepackage{algpseudocode}
\usepackage{graphicx}
\usepackage{multirow}
\usepackage{multicol}
\usepackage{booktabs}
\usepackage{tabularx}
\usepackage{todonotes}
\usepackage{arydshln}

\title{Garbage In, Reasoning Out? \\ Why Benchmark Scores are Unreliable and What to Do About It}

\author{Seyed Mahed Mousavi\textsuperscript{\textdagger}, Edoardo Cecchinato\textsuperscript{\textdagger *}, Lucia Horníková\textsuperscript{\textdaggerdbl}\thanks{Equal Contribution}, Giuseppe Riccardi\textsuperscript{\textdagger}\\
      \textsuperscript{\textdaggerdbl} Masaryk University, Czech Republic \\
      \textsuperscript{\textdagger}   Signals and Interactive Systems Lab, University of Trento, Italy \\
        \texttt{ \{mahed.mousavi,giuseppe.riccardi\}@unitn.it}}

\begin{document}
\maketitle
\begin{abstract}
We conduct a systematic audit of three widely used reasoning benchmarks, SocialIQa, FauxPas-EAI, and ToMi, and uncover pervasive flaws in both benchmark items and evaluation methodology. Using five LLMs (GPT-{3, 3.5, 4, o1}, and LLaMA 3.1) as diagnostic tools, we identify structural, semantic, and pragmatic issues in benchmark design (e.g., duplicated items, ambiguous wording, and implausible answers), as well as scoring procedures that prioritize output form over reasoning process. Through systematic human annotation and re-evaluation on cleaned benchmark subsets, we find that model scores often improve not due to due to erratic surface wording variations and not to improved reasoning. Infact, further analyses show that model performance is highly sensitive to minor input variations such as context availability and phrasing, revealing that high scores may reflect alignment with format-specific cues rather than consistent inference based on the input. These findings challenge the validity of current benchmark-based claims about reasoning in LLMs, and highlight the need for evaluation protocols that assess reasoning as a process of drawing inference from available information, rather than as static output selection. We release audited data and evaluation tools to support more interpretable and diagnostic assessments of model reasoning\footnote{Code and data available at \href{https://github.com/sislab-unitn/GarbageInReasoning}{Repo Link}.}.

\end{abstract}

\section{Introduction}

\begin{figure}[t!]
    \centering
    \includegraphics[width=1\linewidth]{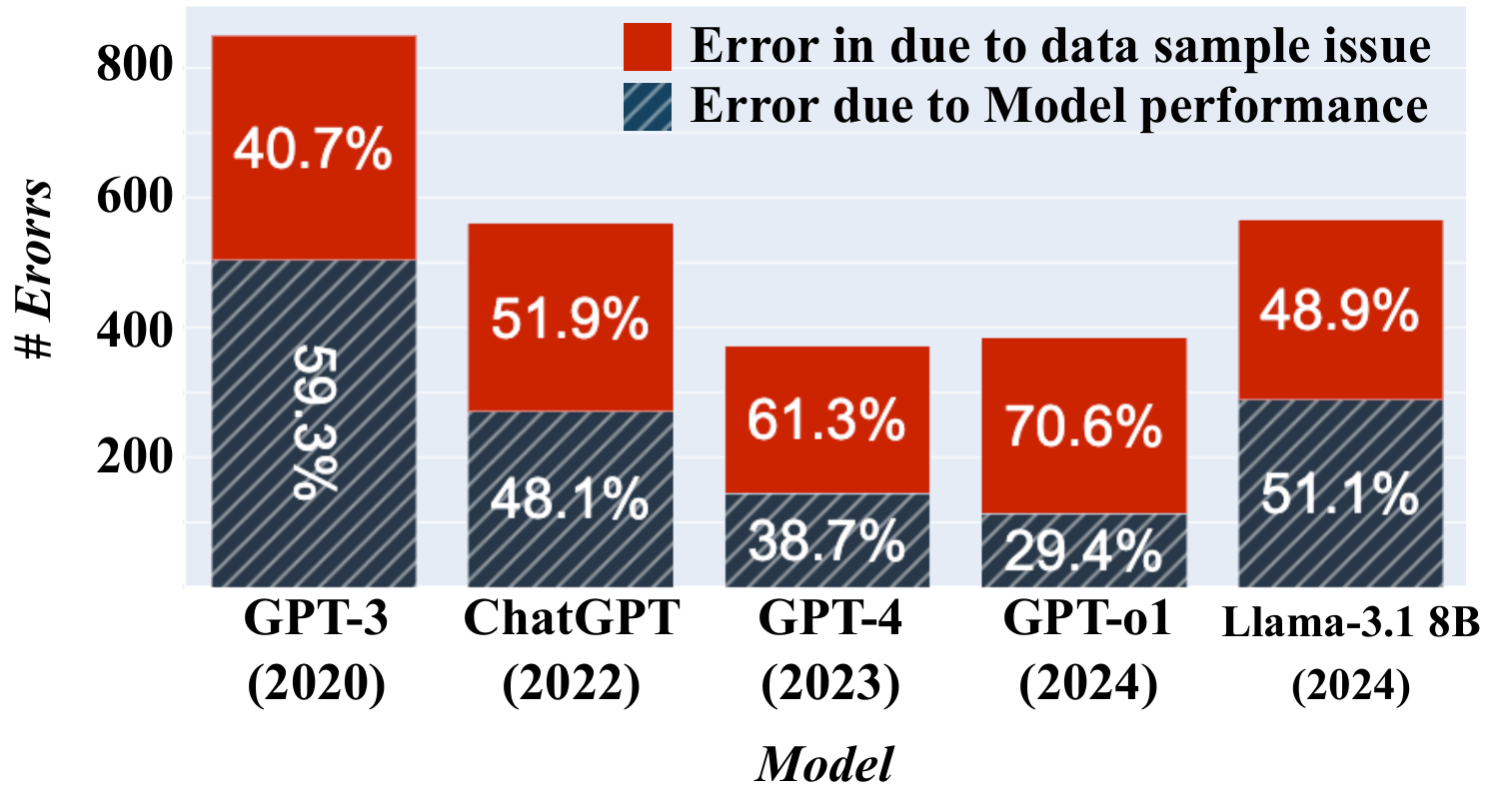}
    \caption{Breakdown of total errors for each model on SocialIQA dev set, divided into \textit{reasoning errors} (errors due to model shortfalls, bottom segment, hatched blue) and \textit{data errors} (errors due to benchmark flaws, top segment, solid red). A substantial portion of errors in each model is consistently due to faulty data.}
    \label{fig:socialqaErrorImpact}
\end{figure}

The state-of-the-art performance of Large language models (LLMs) across downstream tasks has led to a surge of interest in their potential cognitive and reasoning capabilities, including factuality \cite{mousavi2025llms}, common sense reasoning \cite{sap-etal-2019-social,joshi-etal-2025-towards}, and theory of mind \cite{sarangi-etal-2025-decompose}. While several studies embrace LLMs' strong performance on various reasoning benchmarks \cite{webb2023emergent,zhang2025soft,liu2025learn}, other works argue that such performance may be attributed to superficial pattern recognition \cite{shapira-etal-2024-clever}, dataset artifacts \cite{jiang-etal-2024-peek}, or prompt sensitivity \cite{mondorf2024beyond}. This divergence in findings underscores the need to reevaluate existing benchmarks, as it remains unclear whether they reflect the models' ability to reason and draw inferences, or merely mask underlying limitations in how models process and generalize information.

Despite the growing use of benchmarks to evaluate the reasoning abilities of LLMs, closer examination reveals that many of these datasets and evaluation protocols suffer from critical flaws. Much of the current assessment of model reasoning relies on surface-level performance metrics that conflate output accuracy with reasoning capability \cite{mondorf2024beyond}. Models often produce correct outputs through inconsistent or fallacious reasoning processes, revealing a disconnect between prediction and explanation \cite{liu-etal-2024-self-contradictory,jiang-etal-2024-peek}. Recent studies \cite{shapira-etal-2024-clever, holliday-etal-2024-conditional,jiang-etal-2024-peek} further emphasize that benchmark success is often fragile, and highly sensitive to minor rewordings, input context, or task framing. These issues cast doubt on the validity of benchmark-derived claims about LLM reasoning, and they complicate efforts to draw principled conclusions about models’ cognitive capabilities. As a result, both the strengths and weaknesses of current LLMs may be misrepresented, hindering progress toward more robust models.

In this work, we shift the focus from model scores to the validity of the benchmarks and evaluation protocols themselves. Studies have identified isolated issues, such as the prompt sensitivity of ToM evaluations \cite{shapira-etal-2024-clever}, the mismatch between plausibility and gold labels in a subset ($\approx$6\%) of SocialIQa \cite{palta-etal-2024-plausibly}, and self-contradictory reasoning across commonsense datasets \cite{liu-etal-2024-self-contradictory}. However, no prior work offers a systematic, human-verified audit of benchmark quality at scale. 

Our approach complements analyses like \citet{mondorf2024beyond}, who argue for evaluation grounded in reasoning processes rather than outcomes, and \citet{jiang-etal-2024-peek}, who highlight how token-level biases can distort perceived competence. We follow this perspective by defining reasoning as the process of drawing context-sensitive inferences from available information, rather than relying on superficial cues or memorized patterns. By systematically auditing datasets and evaluation protocols, we aim to expose not only where current benchmarks fail, but also how these failures obscure the reasoning shortfalls of models. Our findings advocate for a shift from static outcome-based metrics to evaluations that treat reasoning as a dynamic, explainable process, highlighting both the need for better data and fundamentally new evaluation paradigms.

We examine the reliability of three widely used reasoning benchmarks—SocialIQa \cite{sap-etal-2019-social}, FauxPas-EAI \cite{le-etal-2019-revisiting}, and ToMi \cite{shapira-etal-2023-well}, by evaluating how five language models (GPT-3, ChatGPT (GPT-3.5), GPT-4, GPT-o1, and LLaMA 3.1 8B Instruct) perform under standard prompting and evaluation settings used in prior work \cite{shapira-etal-2024-clever}. Rather than benchmarking the models themselves, we treat them as diagnostic tools to reveal flaws in the datasets and evaluation procedures. We then conduct a structured human audit to systematically identify both data-level and methodological issues, categorize them, and assess their impact on model outputs. Furthermore, we observe that model performance is highly sensitive to surface-level variations, i.e. small changes in phrasing or minor data errors can cause large shifts in performance, exposing a brittle reliance on surface cues rather than reasoning process grounded in context. Our findings reveal that current evaluation setups often the reasoning ability, equating lexical alignment with cognitive competence. Based on these insights, we advocate for a shift toward process-oriented evaluation paradigms that better capture inferential competence. We release audited and annotated data, along with evaluation tools, to support more transparent and interpretable assessments of model reasoning in future research. Our contributions are:

\begin{itemize}
\item We perform a diagnostic assessment of three widely used reasoning benchmarks, SocialIQa, FauxPas-EAI, and ToMi, using five LLMs as tools to uncover data artifacts and methodological inconsistencies.
\item We conduct a large-scale, systematic human audit to identify and categorize dataset-level and evaluation-level flaws, and re-evaluate model performance on curated benchmark subsets to quantify their impact.
\item We expose fundamental limitations of current evaluation practices and argue for a paradigm shift from static output-based metrics toward reasoning-as-process evaluation frameworks.
\item We release the audited data, the annotation materials, and the evaluation tools to support interpretable and diagnostic assessments of reasoning in LLMs.
\end{itemize}

\section{Literature Review}
\citet{webb2023emergent} report that GPT-3 and GPT-4 outperform humans on structured analogical reasoning tasks inspired by cognitive assessments.
\citet{shapira-etal-2024-clever} show that model performance on social reasoning tasks is fragile, driven by shallow heuristics, and correlates with dataset age, suggesting data exposure during pretraining, rather than genuine reasoning ability. \citet{holliday-etal-2024-conditional} find 1) even top-performing models make basic logical errors and inconsistencies; 2) performance correlates with model size, but no model surpasses 90\% accuracy; 3) chain-of-thought prompting improves scores in some settings but fails to eliminate inconsistencies, and in certain cases, introduces new logical failures. \citet{liu-etal-2024-self-contradictory} and \citet{jiang-etal-2024-peek} demonstrate that LLMs frequently produce correct answers using flawed or logically inconsistent reasoning, often relying on superficial cues, and token bias. While the mentioned studies focus mostly on model performance, \cite{palta-etal-2024-plausibly} examine the quality of the benchmarks used for such evaluations. The authors sampled 250 questions ($\approx$6\%) from SocialIQa \cite{sap-etal-2019-social} and CommonsenseQA \cite{talmor-etal-2019-commonsenseqa} as prominent commonsense reasoning benchmarks, and find that over 22\% of samples are problematic. A systematic survey presented by \citet{mondorf2024beyond} argues that evaluation practices often conflate correct answers with sound reasoning, and that data leakage, prompt sensitivity, and shallow metrics inflate perceptions of model competence.

\section{Garbage-In Model Evaluation}
\label{as_is}

We begin by evaluating model performance under standard conditions commonly adopted in prior work \cite{shapira-etal-2024-clever}. Our aim is not to assess the models themselves, but to use their outputs as diagnostic signals. By combining model outputs with human annotation, we identify how flaws in data construction and evaluation protocols distort reported performance and obscure what is actually being measured.

\textbf{SocialIQA} \cite{sap-etal-2019-social} is a crowd-sourced multiple-choice benchmark for evaluating social and emotional commonsense reasoning. Each example consists of a short context, a social inference question, and three answer options, only one of which is correct. As the test set is not publicly released to prevent overfitting and data leakage, we conduct all evaluations on the development set, which contains 1,954 questions. Reported human performance on this subset is 87\%.

\textbf{ToMi} \cite{le-etal-2019-revisiting} is a synthetic benchmark designed to evaluate Theory of Mind (ToM) reasoning in language models. Each story involves multiple agents and object relocations, followed by six questions probing memory, reality, first-order beliefs, and second-order beliefs. ToMi includes randomized story generation, distractor agents, and varied event sequences to reduce exploitable patterns and test belief reasoning more robustly. We evaluate models on a batch of 100 generated stories (600 questions)\footnote{The set is generated using the official code from \citet{le-etal-2019-revisiting}.}.

\begin{figure}[t]
    \centering
    \includegraphics[width=1\linewidth]{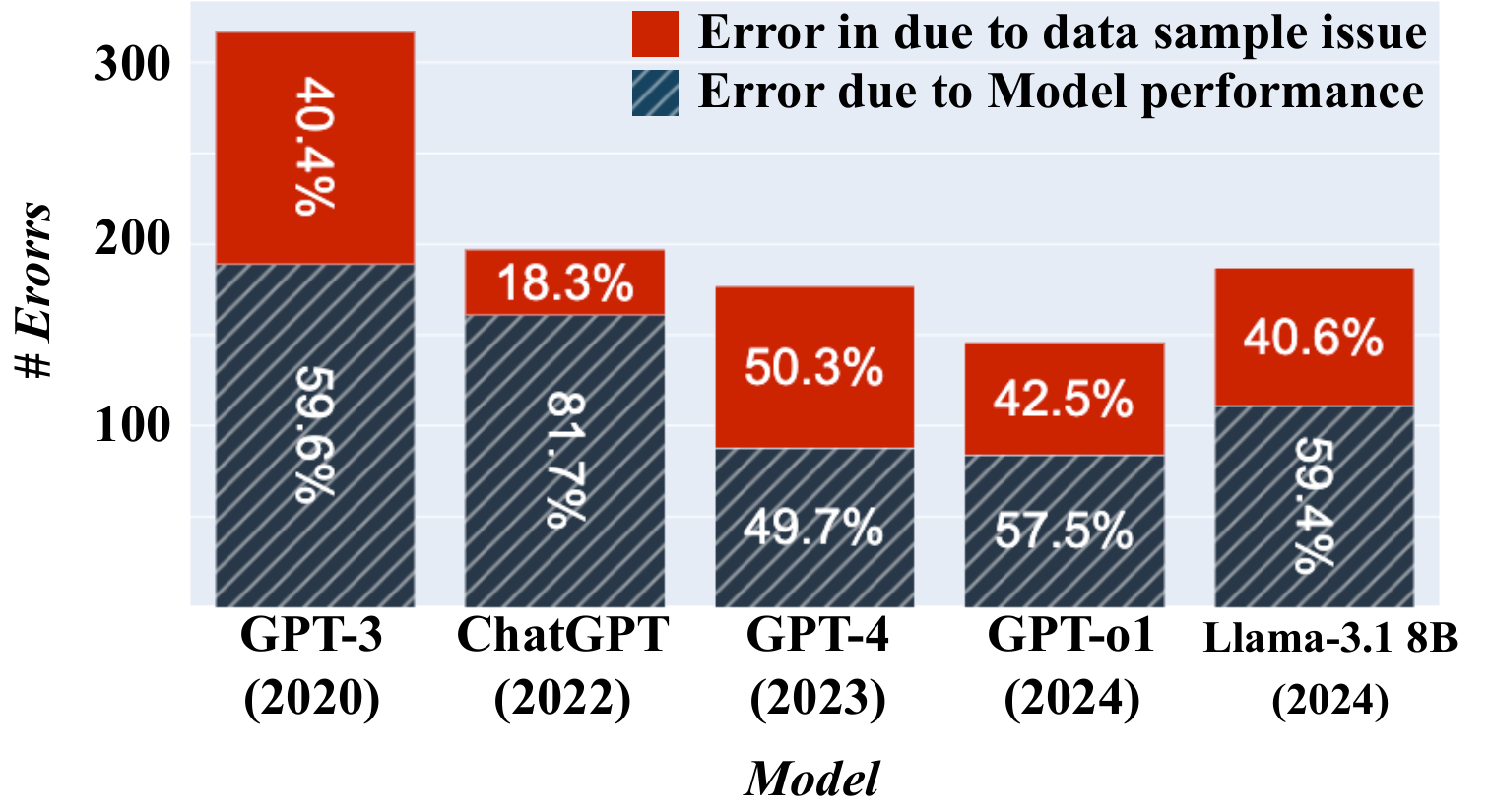}
    \caption{Breakdown of total errors for each model on ToMi, divided into \textit{reasoning errors} (errors due to model shortfalls, bottom segment, hatched blue) and \textit{data errors} (errors due to benchmark flaws, top segment, solid red). Data errors include \textit{Incorrect Ground Truth} and \textit{Answer Granularity Mismatch}.}
    \label{fig:tomiErrorImpact}
\end{figure}

\textbf{FauxPas-EAI} \cite{shapira-etal-2023-well} includes 44 short narrative stories, 22 containing a social faux pas and 22 matched control stories, each followed by four multiple-choice questions: a detection question, a statement identification question, a comprehension check, and a false-belief probe. The benchmark is adapted from a clinical Theory of Mind test \cite{baron1999recognition} designed to assess social reasoning and false-belief understanding. As noted by \citet{shapira-etal-2023-well}, FauxPas-EAI can be viewed as a compositional task that combines SocialIQA and ToMi.

We use five LLMs as diagnostic tools in our benchmark analysis. These include GPT-3\footnote{\href{https://platform.openai.com/docs/models/gpt-base}{davinci-002}} \cite{brown2020language}, an early autoregressive model trained on a broad range of internet text; ChatGPT (GPT-3.5)\footnote{\href{https://platform.openai.com/docs/models/gpt-3-5}{gpt-3.5-turbo-1106}}, a fine-tuned variant optimized for dialogue and instruction-following; GPT-4\footnote{\href{https://platform.openai.com/docs/models/gpt-4-and-gpt-4-turbo}{gpt-4-1106-preview}}, a more advanced model with improved capabilities; GPT-o1\footnote{\href{https://openai.com/o1/}{o1-preview}}, a recent optimized release of GPT-4 with enhanced performance; and, LLaMA 3.1 8B Instruct \footnote{\href{https://ai.meta.com/blog/meta-llama-3-1/}{LLaMA-3.1}}, an open-source instruction-tuned model trained to align with human feedback. Further implementation details are presented in §A.1 and §A.2.

\textbf{Results} As shown in Table~\ref{tab:benchmarks-vs-models}, model performance under standard evaluation settings reproduces known trends across all three benchmarks. Newer models (GPT-4 and GPT-o1) outperform earlier versions such as GPT-3.5 and GPT-3. On SocialIQa, accuracy ranges from 56\% (GPT-3) to 81\% (GPT-4), with intermediate scores from ChatGPT and LLaMA 3.1. A similar pattern holds for ToMi, where performance increases from 47\% to 76\%. On FauxPas-EAI, GPT-4 achieves the highest scores on both question-level and story-level metrics (82\%|39\%), while other models show greater variability. These results align with previous findings \cite{shapira-etal-2024-clever} and confirm that our evaluation setup reproduces previously observed performance trends across models and benchmarks. Full implementation details and prompt templates are provided in our repository.

\section{Benchmark Audit}
We audit benchmark integrity by using LLMs as diagnostic tools; when model predictions appear inconsistent, unstable, or penalized despite valid reasoning process, we investigate whether these behaviors point to underlying flaws in benchmark construction or evaluation. We group these issues into two categories: (1) \textit{Dataset-level issues}, which concern the quality and validity of benchmark items; and (2) \textit{Evaluation-level issues}, which arise from scoring procedures and presentation formats.

\subsection{Data-Related Issues}
We conducted a manual evaluation of all items across the three datasets. Adapting the human evaluation framework proposed by \citet{mousavi-etal-2022-evaluation}, two researchers reviewed each item. The process was supervised by a senior researcher who developed the annotation guidelines and oversaw the process to ensure consistency and annotation quality. Faulty items were identified and categorized into three major dimensions: 1) structural flaws (e.g., duplicated questions, malformed answer options); 2) semantic flaws (e.g., ambiguous wording, internally inconsistent logic); and, 3) pragmatic flaws (e.g., unrealistic scenarios, culturally biased assumptions).

\subsection*{SocialIQA}

\begin{table}[t]
\centering
\small
\begin{tabular}{lcc}
\toprule
\textbf{Error Category} & \textbf{SocialIQA} & \textbf{ToMi} \\
\midrule
 Structural Flaws  & \phantom{-}85 (\phantom{-}4.5\%) & 24\% (\phantom{-}50 Qs) \\
Semantic Flaws     &  361  (18.5\%) & 21\% (126 Qs)\\
Pragmatic Flaws & 130 (\phantom{-}6.7\%)& 45\% (270 Qs) \\
\midrule
\textbf{Total}             & 576 (29.5\%) & - \\
\bottomrule
\end{tabular}
\caption{
Distribution of dataset-level flaws (structural, semantic, and pragmatic) across SocialIQa and ToMi identified through human audit. For SocialIQa, counts and percentages reflect item-level errors across 1,954 development set questions. For ToMi (100 stories, equal to 600 questions), percentages reflect the proportion of stories (and questions) affected within the evaluated batch. The flaws in both benchmarks raise concerns about their reliability for reasoning evaluation.}
\label{tab:socialiqa-errorsdis}
\end{table}

Our audit of the SocialIQa development set revealed that more than 28\% of the items contain data-related flaws that compromise the benchmark’s reliability for evaluating reasoning:

\textbf{I. Structural Flaws ($\approx$4\%)} include surface-level or syntactic issues, such as duplicated entries, repetitive answer options, incomplete prompts, broken syntax, and grammatical errors in the context or answer choices.

\textbf{II. Semantic Flaws ($\approx$18\%)} involve inconsistencies or ambiguities in meaning, such as mismatches in semantic roles (e.g., assigning actions to the wrong participant), violations of temporal logic, referential confusion, and under-specified or contextually vague questions that allow multiple answer options to be equally plausible.

\textbf{III. Pragmatic Flaws ($\approx$7\%)} include answer options that are implausible, socially incoherent, or misaligned with real-world expectations. These include responses that fail to reflect reasonable human motivations, emotional reactions, or common social behavior, as well as answers that are awkwardly phrased or pragmatically unnatural given the context.

\subsection*{ToMi}

While ToMi benefits from the controllability of synthetic generation, this does not guarantee immunity from annotation or construction errors. Our audit of the ToMi benchmark revealed that a significant portion of items contain data-related flaws:

\textbf{I. Structural Flaws ($\approx$0.8\%)} include incorrect ground truth labels that contradict the logical implications of the story, leading to valid model responses being incorrectly marked as wrong, (found in 50 questions affecting 24 stories).

\textbf{II. Semantic Flaws (=21\%)} involve misclassified story types; for example, labeling a false belief as true belief, which breaks the alignment between the intended reasoning task and the question’s logical demands.

\textbf{III. Pragmatic Flaws (=45\%)} include misaligned or inconsistently assigned agent roles, which interfere with perspective-taking tasks (The benchmark defines Agent 0 as the actor who moves the object, and Agent 1 as the other participants). This is particularly damaging in belief and desire questions, where agent identity is central to inferring mental states.

\subsection*{FauxPas-EAI}

We identified data-related issues in 4 (10\%) of the stories. These include: I) \textit{Structural flaws}, such as incorrect labeling of question–answer pairs (e.g., misattributing the gift giver) and awkward or grammatically incorrect phrasing (e.g., “at the kitchen”); II) \textit{Pragmatic flaws}, involving scenarios that fail to clearly constitute a faux pas, as well as questions that hinge on speculative knowledge (e.g., assuming whether a teacher knows a student’s parent), introducing ambiguity that is difficult to resolve in a multiple-choice format. These flawed items span all three story sources; the original clinical set from \citet{baron1999recognition}, the human-authored additions, and the LLM-generated stories by \citet{shapira-etal-2023-well}.

\begin{table*}[t]
\small
\centering
\begin{tabular}{lcccccc}
\toprule
\textbf{Benchmark} & \textbf{\#Samples}&\textbf{GPT-3 } & \textbf{ChatGPT} & \textbf{GPT-4 } & \textbf{GPT-o1} & \textbf{LLaMA 3.1$_{I.}$} \\
\midrule
\textbf{SocialIQA}  & &       &       &       &       &   \\
\hspace{0.1cm}\citet{shapira-etal-2024-clever} & 400 & 19    &   67    & 79      &   -    &    -   \\
\hspace{0.1cm}{\textit{Our Results} (\autoref{as_is})} & 1954& 56    &   71    & 81      &   80    &    71\\  
\hdashline
\noalign{\smallskip} 
\hspace{0.1cm}{Clean Subset (\autoref{re-eval})} & 1378 & 63  &   80    & 90      &   92    &    79   \\
\hline \noalign{\smallskip} 

\textbf{ToMi}           &       &       &       &       &       \\
\hspace{0.1cm}\citet{shapira-etal-2024-clever} & 400 & 39    &   70    & 70      &   -    &    -   \\
\hspace{0.1cm}{\textit{Our Results} (\autoref{as_is})} & 100 & 47    &   67    & 70      &   76    &    68   \\
\hdashline \noalign{\smallskip} 
\hspace{.1cm}{Clean Set (\autoref{re-eval})} & 100 & 69 & 73 & 85 & 86 & 81 \\
\hspace{.60cm}{+ Context} &  & 66 & 84 & 82 & 91 & 79 \\
\hline \noalign{\smallskip}
\textbf{FauxPas-EAI (Q|S)}    &       &       &       &       &       \\
\hspace{0.1cm}\citet{shapira-etal-2024-clever} & 44 & 63 | 14    &   73 | 25    & 74 | 27      &   -    &    -   \\
\hspace{0.1cm}{\textit{Our Results} (\autoref{as_is})} & 44 & 60 | 16    &   75 | 23    & 82 | 39      &   70 | 36    &    61 | 16   \\
\hdashline \noalign{\smallskip} 
\hspace{.1cm}{Clean Subset (\autoref{re-eval})} & 40 & 59 | 13 & 75 | 23 & 84 | 40 & 71 | 40 & 60 | 15 \\
\noalign{\smallskip}
\hspace{.1cm}{Source BreakDown: (\autoref{re-eval})} & & & & &  & \\
\hspace{0.5cm}{\textit{\citet{baron1999recognition}}}&  18 & 60 | 11 & 79 | 22 & 85 | 39 & 67 | 33 & 65 | 22 \\
\hspace{1cm}{+ Context} &  & 71 | 33 & 79 | 39 & 92 | 67 & 74 | 44 & 67 | 22 \\
\hspace{0.5cm}{\textit{\citet{shapira-etal-2023-well} (H.)}}&11  & 54 | 09 & 77 | 27 & 82 | 36 & 75 | 45 & 57 | 09 \\
\hspace{1cm}{+ Context} &  & 59 | 09 & 77 | 18 & 84 | 55 & 80 | 36 & 57 | 27 \\
\hspace{0.5cm}{\textit{\citet{shapira-etal-2023-well} (AI+H.)}} & 11 & 61 | 18 & 66 | 18 & 82 | 45 & 75 | 45 & 57 | 09 \\
\hspace{1cm}{+ Context} &  & 55 | 09 & 73 | 18 & 93 | 82 & 77 | 36 & 66 | 36 \\
\bottomrule
\end{tabular}
\caption{Model performance across the original versions of three benchmarks: SocialIQa, FauxPas-EAI, and ToMi, under different evaluation settings. For each benchmark, we report accuracy (\%) on the original dataset (our results and prior work), the cleaned subset, and versions enhanced with additional context. For \textbf{FauxPas-EAI}, we report question-level (\textbf{Q}) and story-level (\textbf{S}) accuracy, where the latter requires all four questions for a story to be answered correctly. Performance tends to increase with model recency and improves further with cleaned data and contextual prompting.} 
\label{tab:benchmarks-vs-models}
\end{table*}

\subsection{Evaluation-Level Issues}

In addition to dataset-level flaws, we identified several evaluation-level issues that affect how benchmark performance is evaluated and the scores are interpreted. These concern not the content of individual items, but the procedures and assumptions involved in applying the benchmarks and scoring model outputs.

\subsection*{ToMi}

Beyond data-level flaws, our analysis of ToMi reveals several evaluation-level issues that further undermine its reliability as a reasoning benchmark:

\textbf{I. Context Fragmentation} In the literature, the ToMi benchmark is typically applied by presenting each question to the model in isolation, without access to prior questions or answers from the same story. In contrast, human participants process the full narrative and associated questions sequentially, accumulating context throughout. This discrepancy limits the model’s ability to condition its output on earlier events, particularly in second-order belief and desire tasks, where reasoning depends on prior mental state inferences. The fragmented prompting setup thus distorts evaluation by disrupting essential context needed for multi-step inference.

\textbf{II. Semantics-Agnostic Scoring} In ToMi, model outputs are evaluated using strict string matching against gold labels, with no tolerance for lexical variation. This scoring method fails to recognize semantically valid responses. As a result, models may be scored as incorrect despite semantic validity. Conversely, answers that share surface lexical overlap with the gold label may be scored as correct even when they fail to convey the intended meaning. 

\textbf{III. Granularity Mismatch} In $\approx$30\% of responses marked as incorrect, we observed a granularity mismatch between the expected answer and the model’s semantically valid but coarser output. This issue arises from a lack of specificity in the question prompts, which fail to clearly indicate the level of detail required in the answer. For example, when asked to identify the location of an object, expecting the object container as the valid response, models would sometimes respond with the correct room in which the container (and thus the object) is located. While such answers are semantically valid, they are nonetheless penalized.

\subsection*{FauxPas-EAI}
Besides minor data-level issues, we identified several evaluation-level problems that affect how performance is measured on FauxPas-EAI:

\textbf{I. Context Fragmentation} As with ToMi, models are prompted one question at a time in FauxPas-EAI, without access to prior questions or previous answers. In contrast, human evaluators process the full story and all four associated questions as a single, coherent unit. This discrepancy creates a context asymmetry, particularly on identification questions, where reasoning depends on earlier responses. The lack of contextual continuity limits the model’s ability to build and track social inference over multiple steps.

\textbf{II. Semantics-Agnostic Scoring} FauxPas-EAI uses Levenshtein distance \cite{levenshtein1966binary} to score model outputs based on string similarity. This scoring approach rewards lexical alignment over interpretive depth, weakening the benchmark’s diagnostic value. It results in paraphrased or logically correct responses being marked incorrect, while answers that share surface-form overlap with the gold label may be scored as correct even when they fail to reflect valid reasoning. 

\textbf{III. Heterogeneous Story Composition} FauxPas-EAI combines three qualitatively distinct story sources; 20 clinical samples from \citet{baron1999recognition}, 12 human-authored stories, and 12 LLM-generated stories validated post hoc \cite{shapira-etal-2023-well}. While treated as a unified benchmark in the literature, these subsets differ in narrative complexity and reasoning demands. The original clinical stories rely on implicit social knowledge and belief modeling, while the newer stories are more explicit and structurally simplified \cite{shapira-etal-2023-well}. Merging these stories into a single evaluation set can inflate model scores in misleading ways.

\subsection{Analysis and Implications}

Our assessment reveals pervasive flaws across all three benchmarks, spanning both dataset-  and evaluation-levels. Structural, semantic, and pragmatic flaws (Table~\ref{tab:socialiqa-errorsdis}) compromise the integrity of benchmark items, while semantic-agnostic scoring protocols and context-fragmented prompting fail to reflect the reasoning capability  they aim to measure. These errors inject noise into evaluation and distort conclusions. As a result, models may be penalized for appropriate reasoning or rewarded for token overlap. Figures \ref{fig:socialqaErrorImpact} and~\ref{fig:tomiErrorImpact} illustrate that, in both SocialIQa and ToMi, a large share of model “errors” are attributable not to model limitations, but to flaws in benchmark design.

Benchmarks relying on multiple-choice formats (SocialIQA) or pre-defined task schemas (ToMi, FauxPas-EAI) reduce reasoning to surface-level selection. When tasks are framed as choosing from a fixed set of options without requiring explanation or justification, correct reasoning becomes indistinguishable from shallow token overlap. In such settings, correct answers can arise from token-level heuristics rather than conceptual understanding. These setups tend to reward surface-level alignment over valid reasoning processes and equate output correctness with reasoning competence. More broadly, they reflect a deeper limitation of current benchmark design: \textbf{benchmarks based on pre-defined schemas and token-matching metrics fail to capture the flexible, contextual nature of reasoning}, and risk conflating format adherence with cognitive ability.

\begin{figure}[t]
    \centering
    \includegraphics[width=1\linewidth]{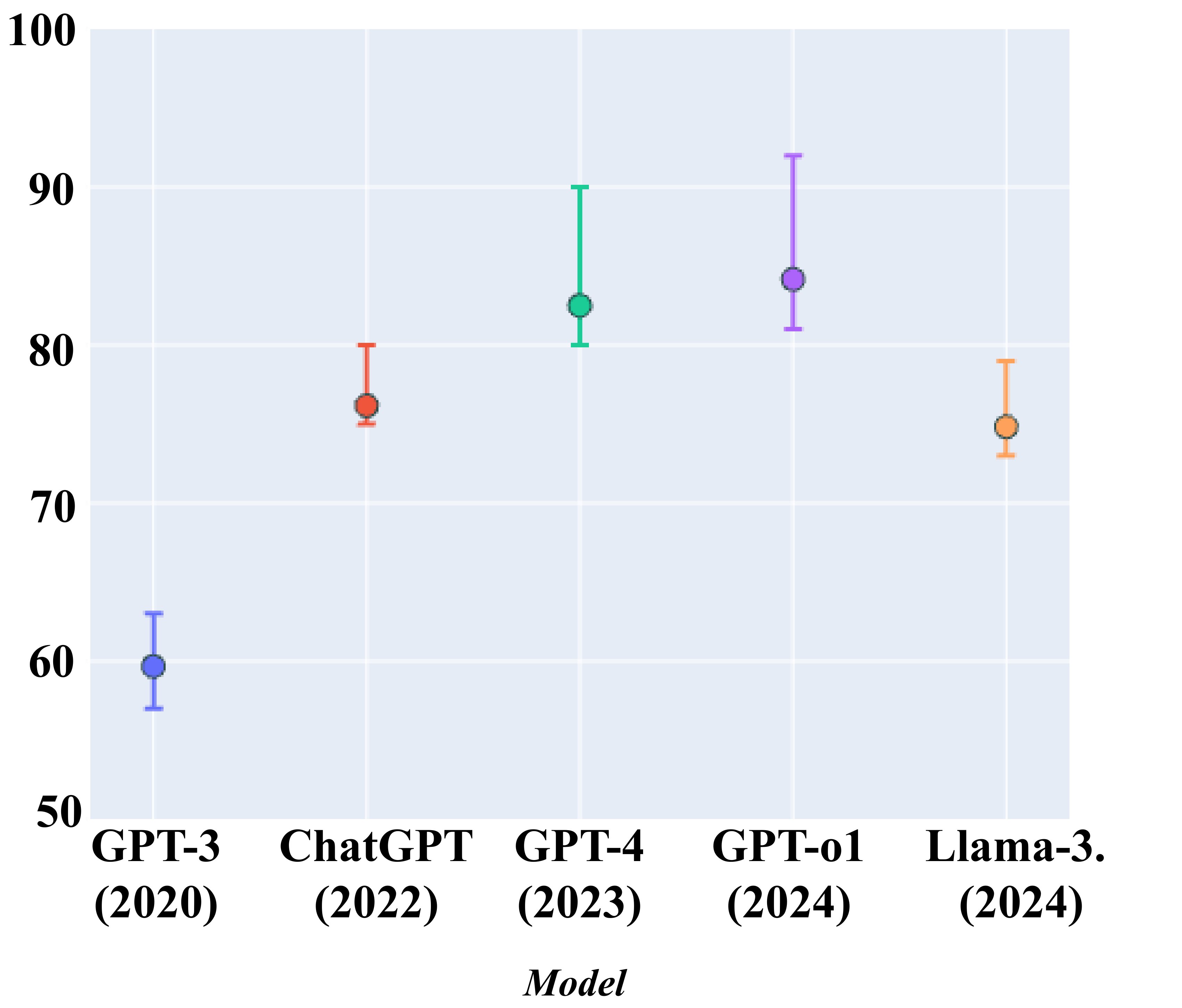}
    \caption{Performance on SocialIQa (clean subset) and five semantically equivalent rephrasings per item. Despite no change in content, all models exhibit notable score fluctuations, revealing sensitivity to surface-level variation and the limitation of the benchmark to assess robust reasoning.}
    \label{fig:fluct}
\end{figure}

\section{High Scores Conceal Reasoning Gaps}

\label{re-eval}

To better understand what current benchmark scores reflect, we examine model performance under modified evaluation settings that address flaws identified in our audit. Our goal is to assess how reported performance is shaped by data quality, context availability, input phrasing, and scoring methodology. We analyze four dimensions: cleaned datasets (A), context-aware prompting (B), semantically equivalent rephrasings (C), and meaning-aware scoring using LLMs (D). This analysis probes the stability of model predictions and the robustness of benchmark scores to input-level noise, including annotation inconsistencies, missing context, and superficial lexical variation. 

\textbf{A. Evaluation on Cleaned Data} We investigate how performance scores vary when input-noise is minimized. We filtered out flawed items from SocialIQA and FauxPas-EAI, corrected labeling and agent-role inconsistencies in ToMi, adjusted scoring procedures to account for granularity mismatches, and manually reviewed all model responses. As shown in Table~\ref{tab:benchmarks-vs-models}, all models exhibit improved accuracy on the cleaned datasets. For example, GPT-o1 achieves 92\% on the cleaned SocialIQA subset (up from 80\%), and GPT-4 improves to 85\% on ToMi. FauxPas-EAI also shows modest gains after data cleaning, though performance remains variable. Nevertheless, rather than enhanced reasoning ability, \textbf{these gains highlight how sensitive model performance is to input-noise, annotation quality, and benchmark design}.  We also examine performance across the three story sources in FauxPas-EAI (clinical, human-authored, and LLM-generated), and observe notable variation across subsets. This variation suggests that differences in narrative structure and complexity, rather than reasoning competence, may account for observed model performance.

\textbf{B. Context-Aware Evaluation} We evaluate models in a context-aware setting, where they are exposed to the full sequence of story questions. This allows to accumulate context across turns, particularly important for tasks involving belief tracking or multi-step inference. As shown in Table~\ref{tab:benchmarks-vs-models}, performance improves further under this setup for most models. Nevertheless, the performance of a few models drop slightly in across benchmarks, suggesting that not all models benefit equally from additional context. These results confirm that isolated question prompting underestimates model performance on tasks requiring sequential social reasoning. However, \textbf{higher scores in context-aware settings do not signal improved reasoning, but rather expose the inadequacy of fragmented evaluation setups}.

\textbf{C. Format Sensitivity} To investigate the impact of input format on model performance, we generated five alternative phrasings for each item in the cleaned SocialIQa subset using GPT-4o. A random 15\% of the generated items were manually validated to ensure semantic equivalence and linguistic correctness. We then re-evaluated model performance on these rephrasings. Despite no change in content, model scores varied considerably, Figure \ref{fig:fluct}. These findings reveal that what performance scores on static benchmarks represent is not reasoning process, but a fragile reliance on specific lexical patterns and token-level cues.  The observed volatility underscores that even minimal input variation can destabilize model performance, exposing a brittleness inconsistent with inferential capabilities. \textbf{Rather than assessing models' ability to draw inferences from the input context, current static benchmarks reflect sensitivity to surface-level cues and variations in phrasing.}

\textbf{D. LLM-as-a-Judge} In this evaluation, we relied on human annotations to mitigate the limitations of semantics-agnostic scoring. Nevertheless, we explored the use of LLM-as-a-judge \cite{ho2025llm, bavaresco2024llms} as a scalable cost-effective surrogate for meaning-aware evaluation. We assessed GPT-4o in evaluating the semantic validity of model predictions given the question and gold reference. On FauxPas-EAI items, GPT-4o achieved near-perfect agreement with human annotations (Cohen’s $\kappa = 0.89$), outperforming Levenshtein distance ($\kappa = 0.77$). These findings suggest that LLMs can provide an alternative to rigid string-matching metrics. However, we emphasize that \textbf{LLM-as-a-judge should be treated as a diagnostic aid, not a substitute for rigorous human evaluation}.

\section{Discussion}
As demonstrated in Section \ref{re-eval}, high benchmark scores often collapse under input variation or when superficial cues are removed, revealing a persistent gap between what benchmark scores reflect and the actual reasoning processes exhibited by models. Once flawed data is corrected and phrasing is controlled, models struggle to produce stable, contextually grounded inferences. This suggests that current benchmarks do not reliably assess reasoning as inference from input content, but instead reward lexical conformity and alignment with format-specific cues.

\textbf{Static Benchmarks Conflate Accuracy with Reasoning.} Current static benchmarks reduce reasoning to static tasks evaluated through string matching or token-level overlap, failing to reflect reasoning as a dynamic, contextual, and inferential process. Pre-defined formats and static schema encourage pattern matching and fluency over correct reasoning process. Our audit and prior work \cite{jiang-etal-2024-peek, liu-etal-2024-self-contradictory} show that models often produce correct answers while offering incoherent or contradictory rationales. Even GPT-o1, which scores 92\% on the cleaned SocialIQa subset surpassing human accuracy, exhibits brittle behavior when confronted with minor rephrasings or alternative phrasings of the same question. Such instability (Section~\ref{re-eval}) reveals that what benchmarks reward is not inference, but format sensitivity. \textit{A model can achieve perfect accuracy and still fail to reason.}

\textbf{Benchmarks Create the Illusion of Progress.} Observed high scores are frequently driven by overfitting to benchmark artifacts, not gains in reasoning ability. Many widely-used datasets suffer from contamination, outdated task designs, or structural inconsistencies. This inflates performance through memorization or exposure, rather than generalization. Moreover, models that cannot maintain logical consistency and frequently contradict their own outputs when asked to explain themselves, lack the required coherence for reasoner models. Their success is not evidence of reasoning but of matching familiar patterns and memorization. Without contextual robustness or internal consistency, high scores serve as a misleading proxy for cognitive competence.

\textbf{Toward Process-Aware Evaluation.} The shortcomings we identify reflect not just model limitations, but deeper flaws in how reasoning is currently evaluated. Current benchmarks reward surface-level alignment over inferential capabilities, emphasizing outcomes rather than the reasoning processes that produce them. More meaningful alternatives include structured rationale evaluation, contradiction detection, counterfactual testing, and interactive formats such as adaptive diagnostics \citep{mondorf2024beyond}. These approaches prioritize reasoning as a process rather than an outcome, and offer a more diagnostic view of model competence. By using current benchmarks, we risk mistaking statistical fluency for understanding, and building models that appear intelligent, but are not.

\section{Conclusion}
We conducted a detailed audit of widely used reasoning benchmarks and evaluation practices to assess what current methods truly measure in LLMs. Our findings reveal that benchmark performance is frequently shaped by flawed data, fragile evaluation protocols, and a reliance on surface-level cues. Even when models achieve high scores, their outputs often reflect surface lexical alignment rather than input grounded reasoning process. In particular, we find that performance can degrade substantially under semantically equivalent rephrasings or minor data imperfections, highlighting a deep sensitivity to input noise that undermines reliability.

We argue that meaningful progress in developing and evaluating reasoning capabilities requires methods that treat reasoning as a contextual, coherent, and explainable process. To move forward, we must shift away from static, form-driven metrics toward process-level evaluation, tracking reasoning steps, identifying contradictions, and testing robustness across rephrasings and contexts. Without such a shift, benchmark success will continue to misrepresent what LLMs can actually do, and claims of reasoning competence will remain on shaky ground.

\section*{Limitations}
Our study focuses only social reasoning and theory-of-mind tasks, leaving logical and mathematical reasoning benchmark unexplored. Our rephrasing experiments rely on a single backbone model, which may limit the generality and diversity of our format sensitivity findings. While our human audit is systematic, its scale is resouce-intensive. Extending this diagnostic approach to more diverse reasoning tasks, larger datasets, and multilingual settings remains important future work.



\bibliography{custom}

\newpage
\onecolumn

\appendix

\appendix
\section{Appendix}

\subsection{Implementation Details}
For every benchmark, we ran the experiments using five different models: GPT-3\footnote{\href{https://platform.openai.com/docs/models/gpt-base}{davinci-002}}, ChatGPT (GPT-3.5)\footnote{\href{https://platform.openai.com/docs/models/gpt-3-5}{gpt-3.5-turbo-1106}}, GPT-4\footnote{\href{https://platform.openai.com/docs/models/gpt-4-and-gpt-4-turbo}{gpt-4-1106-preview}}, GPT-o1\footnote{\href{https://openai.com/o1/}{o1-preview}}, and LLaMA 3.1 8B Instruct\footnote{\href{https://ai.meta.com/blog/meta-llama-3-1/}{LLaMA-3.1}}. We used the official OpenAI APIs\footnote{\href{https://platform.openai.com/docs/overview}{OpenAI developer platform}} for the first four and the HuggingFace APIs for LLaMA 3.1\footnote{\href{https://huggingface.co/meta-llama/Llama-3.1-8B-Instruct}{meta-llama/Llama-3.1-8B-Instruct}}.
We set the temperature of all the models to 0 for reproducibility reasons, with the only exception of GPT-o1 for which is set to 1 by default.
We queried every model sequentially for every data entry, and following we present the number of queries for each benchmark and related tasks.

\textbf{SocialIQA}\\
We ran models evaluation for all the development set (1954 items), that for all the five models implies 9770 calls to the APIs. After this step, we generated five rephrasings for every data entry of the cleaned version of the benchmark (1378 items). To do so we queried GPT-4o\footnote{\href{https://platform.openai.com/docs/models/gpt-4o}{gpt-4o-2024-08-06}} 6890 times, using a temperature value of 1. Finally, we ran again all the initial models on the rephrasings (6890 per model), adding a number of 34,450 API calls.
At the end of the experiment, for SocialIQA we obtained a total 51,110 calls.

\textbf{ToMi}\\
For ToMi we generated 100 stories from the original repository\footnote{\href{https://github.com/facebookresearch/ToMi/blob/master/README.md}{facebookresearch/ToMi}}, each containing 6 questions. For all the models, we initially evaluated every question sequentially (3000 API calls), and then providing the conversation history (3000 API calls). The conversation history consists of adding the previously answered questions for the same story, in addition to the context and the current question. In the end, the total number of queries for this benchmark was 6000.

\textbf{FauxPas-EAI}\\
FauxPas-EAI consists of 44 stories, each containing 4 questions (176 questions in total). Like we did for ToMi, we evaluated each model sequentially (880 API calls) and with the conversation history (880 API calls). Then, on the cleaned dataset, containing 40 stories (160 questions), we queried GPT-4o (temperature set to 0) as LLM-as-a-Judge evaluator instead of using Levenshtein distance. Doing this for every model predictions we executed 800 additional API calls. These LLM annotations were then compared with the humans ones using the Cohen’s $\kappa$ value, implemented in this repository\footnote{\href{https://github.com/sislab-unitn/Human-Evaluation-Protocol/blob/main/scripts/iaa_metrics.py}{sislab-unitn/Human-Evaluation-Protocol}}. In the end, the final total number of queries for FauxPas-EAI was 2560.

\subsection{Prompt Examples}

Table~\ref{tab:evaluation-prompts} and ~\ref{tab:additional-prompts} contain all the prompts used to run our experiments. The first one contains the benchmarks evaluation prompts, while the second one contains the prompts used for the SocialIQA rephrasing task and the FauxPas-EAI LLM-as-a-Judge evaluation.

\begin{table*}[t]
\small
\centering
\begin{tabular}{p{3cm} p{11cm}}
\toprule
\textbf{Benchmark} & \textbf{Prompt} \\
\midrule
\textbf{SocialIQA}  & \\
\hspace{0.1cm} Sequential & \textit{Context: \textbraceleft context\textbraceright\newline Question: \textbraceleft question\textbraceright\newline Based on the context, which option is the best answer to the question?\newline Option 1: \textbraceleft option1\textbraceright\newline Option 2: \textbraceleft option2\textbraceright\newline Option 3: \textbraceleft option3\textbraceright\newline Answer with ``1'', ``2'' or ``3'', without explanations. In case of doubt, answer according to the most probable answer.} \\ 
\hline \noalign{\smallskip} 

\textbf{ToMi} &  \\
\hspace{0.1cm}Sequential & \textit{Context: \textbraceleft context\textbraceright \newline Question: \textbraceleft question\textbraceright Answer with one word only, without explanations. Answer: }\\
\hdashline \noalign{\smallskip} 
\hspace{0.1cm}Conv-history  & \textit{[\textbraceleft"role": "user", "content": \textbraceleft prompt1\textbraceright\textbraceright\newline
  \textbraceleft"role": "assistant", "content": \textbraceleft model\_answer1\textbraceright\textbraceright,\newline
  \textbraceleft"role": "user", "content": \textbraceleft prompt2\textbraceright\textbraceright,\newline
  \textbraceleft"role": "assistant", "content": \textbraceleft model\_answer2\textbraceright\textbraceright,\newline
  \dots\newline
  \textbraceleft"role": "user", "content": \textbraceleft prompt6\textbraceright\textbraceright]} \\
\hline \noalign{\smallskip}
\textbf{FauxPas-EAI} & \\
\hspace{0.1cm}Sequential & \textit{\textbraceleft story\textbraceright \newline \textbraceleft question\textbraceright \newline Answer:} \\ 
\hdashline \noalign{\smallskip} 
\hspace{0.1cm}Conv-history  & \textit{[\textbraceleft"role": "user", "content": \textbraceleft prompt1\textbraceright\textbraceright\newline
  \textbraceleft"role": "assistant", "content": \textbraceleft model\_answer1\textbraceright\textbraceright,\newline
  \textbraceleft"role": "user", "content": \textbraceleft prompt2\textbraceright\textbraceright,\newline
  \textbraceleft"role": "assistant", "content": \textbraceleft model\_answer2\textbraceright\textbraceright,\newline
  \dots\newline
  \textbraceleft"role": "user", "content": \textbraceleft prompt4\textbraceright\textbraceright]} \\
\bottomrule
\end{tabular}
\caption{Prompts used for the benchmarks evaluation, both sequentially and with conversation history. In SocialIQA, for GPT-3 and LLaMA 3.1 we slightly changed the prompt structure, adding some guideline for the specific output formatting.} 
\label{tab:evaluation-prompts}
\end{table*}

\begin{table*}[t]
\small
\centering
\begin{tabular}{p{3cm} p{11cm}}
\toprule
\textbf{Task} & \textbf{Prompt} \\
\midrule
\textbf{SocialIQA rephrasing}  & \textit{Your job is to generate 5 different rephrasings of the given item. For each rephrasing, rephrase the context, question and possible answers while maintaining their semantics and structure.\newline
  \#\# Response Format:\newline
  Each rephrasing should be separated by "---REPHRASING---" and follow this exact format:\newline
  REPHRASING 1:\newline
  Context: [rephrased context]\newline
  Question: [rephrased question]\newline
  Answer A: [rephrased answer A]\newline
  Answer B: [rephrased answer B]\newline
  Answer C: [rephrased answer C]\newline
  ---REPHRASING---\newline
  REPHRASING 2:\newline
  [same format as above]\newline
  And so on for all 5 rephrasings.\newline
  \#\# Original Item to Rephrase:\newline
  Context: \textbraceleft context\textbraceright\newline
  Question: \textbraceleft question\textbraceright\newline
  Answer A: \textbraceleft answerA\textbraceright\newline
  Answer B: \textbraceleft answerB\textbraceright\newline
  Answer C: \textbraceleft answerC\textbraceright} \\ 
\hline \noalign{\smallskip} 

\textbf{LLM-as-a-Judge (FauxPas-EAI)} &  \textit{Your job is to evaluate a predicted answer by comparing it against the gold answer and the given question. You may refer to the provided context if needed.\newline
  \#\# Grading Criteria:\newline
  * 1: The predicted answer matches the gold answer or is a valid alternative (e.g.,different but correct ways of writing a name).\newline
  * 0: The predicted answer is wrong or does not align with the gold answer.\newline
  * In some ambiguous cases, where it is unclear whether the predicted answer is correct or not, please refer to the provided context and use it as the final source for making your judgment.\newline
  \#\# Response Format:\newline
  Please answer with ONLY '1' or '0'.\newline
  \#\# Here is your task:\newline
  Context: \textbraceleft context\textbraceright\newline
  Question: \textbraceleft question\textbraceright\newline
  Gold Answer: \textbraceleft original\_answer\textbraceright\newline
  Predicted Answer: \textbraceleft predicted\_answer\textbraceright} \\ 
\bottomrule
\end{tabular}
\caption{Prompts used for additional evaluation tasks. The rephrasing prompt was used to generate five rephrased versions of every item in the cleaned version of SocialIQA, while the LLM-as-a-judge one to run automatic evaluation for FauxPas-EAI, instead of the originally used Levenshtein distance.} 
\label{tab:additional-prompts}
\end{table*}

\end{document}